\documentclass[11pt,letterpaper]{article}
\usepackage{naaclhlt2013}
\usepackage{times}
\usepackage{latexsym}
\usepackage{amsmath}
\usepackage{multirow}
\usepackage{url}
\usepackage{color}
\usepackage{graphicx}

\usepackage{algorithmic, amssymb, amsmath}   
\usepackage{algorithm}
\usepackage{latexsym}
\usepackage[T5,T1]{fontenc}

\title{Creating Reverse Bilingual Dictionaries}

\author{Khang Nhut Lam\\
Department of Computer Science\\
University of Colorado\\
Colorado Springs, USA\\
{\tt klam2@uccs.edu}
\And
Jugal Kalita\\
Department of Computer Science\\
University of Colorado\\
Colorado Springs, USA\\
{\tt jkalita@uccs.edu}}

\begin{document}
\maketitle
\begin{abstract}
Bilingual dictionaries are expensive resources and not many are available when one of the languages is resource-poor. In this paper, we propose algorithms for creation of new reverse bilingual dictionaries from existing bilingual dictionaries in which English is one of the two languages. Our algorithms exploit the similarity between word-concept pairs using the English Wordnet to produce reverse dictionary entries. Since our algorithms rely on available bilingual dictionaries, they are applicable to any bilingual dictionary as long as one of the two languages has Wordnet  type lexical ontology.
\end{abstract}

\section{Introduction}
The Ethnologue organization\footnote{http://www.ethnologue.com/} lists 6,809 distinct languages in the world, most of which are resource-poor.  Most existing online bilingual dictionaries are between two  resource-rich languages (e.g., English, Spanish, French or German) or between a resource-rich language and a resource-poor language.  There are languages for which we are lucky to find a single bilingual dictionary online. For example, the University of Chicago hosts  bilingual dictionaries from 29 Southeast Asian languages\footnote{http://dsal.uchicago.edu/dictionaries/list.html}, but many of these languages have only one bilingual dictionary online.

Existing algorithms for creating new bilingual dictionaries use intermediate languages or intermediate dictionaries to find chains of words with the same meaning. For example,~\cite{Gollins:2001} use lexical triangulation to translate in parallel across multiple intermediate languages and fuse the results. They query several existing dictionaries and then merge results  to maximize accuracy. They use four pivot languages, German, Spanish, Dutch and Italian, as intermediate languages.  
Another existing approach for creating bilingual dictionaries is using probabilistic inference~\cite{Mausam:2010}. They organize dictionaries in a graph topology and use random walks and probabilistic graph sampling.~\cite{Shaw:2011} propose a set of algorithms to create a reverse dictionary in the context of single language by using converse mapping.  In particular, given an English-English dictionary, they attempt to find the original words or terms given a synonymous word or phrase describing the meaning of a word. 

The goal of this research is to study the feasibility of creating a reverse dictionary by using only one existing dictionary and Wordnet lexical ontology.  For example, given a Karbi\footnote{Karbi is an endangered language spoken by 492,000 people (2007 Ethnologue data) in Northeast India, ISO 639-3 code AJZ. ISO 693-3 code for English is ENG.}-English dictionary, we will construct an ENG-AJZ dictionary.
The remainder of this paper is organized as follows.  In Section 2, we discuss the nature of bilingual dictionaries. Section 3 describes the algorithms we propose to create new bilingual dictionaries from existing dictionaries. Results of our experiments are presented in Section 4. Section 5 concludes the paper.

\section {Existing Online Bilingual Dictionaries}  
Powerful online translators developed by  Google and Bing  provide pairwise translations (including for individual words) for  65 and 40 languages, respectively. Wiktionary, a  dictionary created by volunteers, supports  over 170 languages.
We find a large number of bilingual dictionaries at PanLex\footnote{http://panlex.org/} including an ENG-Hindi\footnote{ISO 693-3 code HIN} and a Vietnamese\footnote{ISO 693-3 code VIE}-ENG dictionary.
The University of Chicago has a number of bilingual dictionaries for South Asian languages. Xobdo\footnote{http://www.xobdo.org/} has a number of dictionaries, focused on Northeast India.  

We classify the many freely available dictionaries into three main kinds.

\vspace*{-.125in}

\begin{itemize}
\setlength{\itemsep}{-2pt}
\item Word to word dictionaries: These are dictionaries that translate  one word in one language to one word or a phrase in another language. An example is an ENG-HIN dictionary  at Panlex. 

\item Definition dictionaries: One word in one language has one or more meanings in the second language. It also may have pronunciation, parts of speech, synonyms and examples. An example is the VIE-ENG dictionary, also at Panlex.

\item One language dictionaries: A dictionary of this kind is found at dictionary.com. 
\end{itemize}
\vspace*{-.08in}
We have examined several hundred online dictionaries and found that they occur in many different formats. Extracting  information from these dictionaries is arduous. We have experimented with five existing bilingual dictionaries: VIE-ENG, ENG-HIN, and a dictionary supported by Xobdo with 4 languages: Assamese\footnote{Assamese is an Indo-European language spoken by about 30 million people, but it is resource-poor, ISO 693-3 code ASM.}, ENG, AJZ, and Dimasa\footnote{Dimasa is another endangered language from Northeast India, spoken by about 115,000 people, ISO 693-3 code DIS.}. We consider the last one to be a collection of 3 bilingual dictionaries: ASM-ENG, AJZ-ENG, and DIS-ENG. We choose these languages since one of our goals is to work with resource-poor languages to enhance the quantity and quality of resources available. 

\section{Proposed Solution Approach}
A dictionary entry, called {\em LexicalEntry}, is a 2-tuple {\em $<$LexicalUnit, Definition$>$}. A {\em  LexicalUnit} is a word or a phrase being defined, also called {\em definiendum}~\cite{Landau:1984}.  A list of entries sorted by the LexicalUnit is called a {\em lexicon} or a {\em dictionary}. Given a {\em LexicalUnit}, the {\em Definition} associated with it usually contains its class and pronunciation, its   {\em meaning}, and possibly additional information. The  meaning  associated with it can have several {\em Sense}s. A {\em Sense} is a discrete representation of a single aspect of the meaning of a word. Thus, a dictionary entry is  of the form {\em \textless LexicalUnit, ${Sense}_1$, ${Sense}_2,\cdots$\textgreater}. 

In  this section, we propose a series of  algorithms, each one of which  automatically creates a reverse dictionary, or $ReverseDictionary$, from a dictionary that translates a word in language $L_1$ to a word or phrase in language $L_2$.  
We require that at least one of two these languages has a Wordnet type lexical ontology~\cite{Miller:1995}. Our algorithms are used to create reverse dictionaries from them at various levels of accuracy and sophistication. 
 
\subsection{Direct Reversal (DR)}
The existing dictionary has alphabetically sorted {\em LexicalUnit}s in $L_1$ and each of them has one or more \emph{Sense}s in $L_2$. To create {\em ReverseDictionary}, we simply take every pair \textless$LexicalUnit, Sense$\textgreater \; in {\em SourceDictionary} and swap the positions of the two. 
\begin{algorithm}[h]
\caption{DR Algorithm}
\begin{algorithmic}
\STATE $ReverseDictionary$ := $\phi$
\FORALL{${LexicalEntry}_i\in$ $SourceDictionary$}
\FORALL {${Sense}_j \in {LexicalEntry}_i$}
\STATE Add tuple \textless ${Sense}_j$, ${LexicalEntry}_i.LexicalUnit$\textgreater\; to $ReverseDictionary$
\ENDFOR 
\ENDFOR
\end{algorithmic}
\end{algorithm}

This is a baseline algorithm so that we can compare improvements as we create new algorithms. If in our input dictionary, the sense definitions are mostly single words, and occasionally a simple phrase, even such a simple algorithm gives  fairly good results. In case there are long or complex phrases in senses, we  skip them. The approach is easy to implement, and produces a high-accuracy $ReverseDictionary$. However, the number of entries in the created dictionaries are limited because this algorithm just swaps the positions of \emph{LexicalUnit} and \emph{Sense} of each entry in the $SourceDictionary$  and does not have any method to find the additional words having the same meanings.
 
\subsection{Direct Reversal with Distance (DRwD)}
To increase the number of entries in the output dictionary, we  compute the distance between words in the Wordnet hierarchy. For example, the words "hasta-lipi" and "likhavat" in HIN have the meanings "handwriting" and "script", respectively. The distance between "handwriting" and "script" in Wordnet hierarchy is 0.0, so that "handwriting" and "script" likely have the same meaning. Thus, each of "hasta-lipi" and "likhavat" should have both meanings "handwriting" and "script". This approach helps us find additional words having the same meanings and  possibly increase the number of lexical entries in the reverse dictionaries.

To create a \emph{ReverseDictionary}, for every $LexicalEntry_i$ in the existing dictionary, we find all $LexicalEntry_j, i\neq j$ with distance to $LexicalEntry_i$ equal to or smaller than a threshold $\alpha$. As results, we have new pairs of entries  \textless$LexicalEntry_i.LexicalUnit$, $LexicalEntry_j.Sense$\textgreater\;; then we swap positions in the two-tuples, and add them into the \emph{ReverseDictionary}. The value of $\alpha$ affects the number of entries and the quality of  created dictionaries. The greater the value of $\alpha$, the larger the number of lexical \emph{entries}, but the smaller the accuracy of the {\em ReverseDictionary}.

The distance between the two \emph{LexicalEntry}s is the distance between the two $LexicalUnit$s if the {\em LexicalUnit}s occur in Wordnet ontology; otherwise, it is the distance between the two $Sense$s. The distance between each phrase pair is the average of the total distances between every word pair in the phrases \cite{Wu:1994}. If the distance between two words or phrases  is 1.00, there is no similarity between these words or phrases, but if they have the same meaning, the distance is 0.00.

We find that a $ReverseDictionary$ created  using the value 0.0 for  $\alpha$ has the highest accuracy. This approach significantly increases the number of entries in the \emph{ReverseDictionary}. However, there is an issue in this approach. For instance, the word "tuhbi" in DIS means "crowded", "compact", "dense", or "packed". Because the distance between the English words "slow" and "dense" in Wordnet  is 0.0, this algorithm concludes that "slow" has the meaning "tuhbi" also, which is wrong.
 \begin{algorithm}
\caption{DRwD Algorithm}
\begin{algorithmic}
\STATE $ReverseDictionar$y := $\phi$
\FORALL {${LexicalEntry}_i\in$ $SourceDictionary$}
\FORALL  {${Sense}_j \in {LexicalEntry}_i$}
\FORALL {${LexicalEntry}_u\in$ $SourceDictionary$}
\FORALL  {${Sense}_v \in {LexicalEntry}_u$}
\IF{$distance$(\textless${LexicalEntry}_i.LexicalUnit$,
${Sense}_j$\textgreater\;,\textless ${LexicalEntry}_u.LexicalUnit$, ${Sense}_v$\textgreater\;) $\leqslant \alpha$}
\STATE Add tuple \textless${Sense}_j$, ${LexicalEntry}_u.LexicalUnit$\textgreater\; to $ReverseDictionary$
\ENDIF 
\ENDFOR 
\ENDFOR 
\ENDFOR 
\ENDFOR 
\end{algorithmic}
\end{algorithm}

\subsection{Direct Reversal with Similarly (DRwS)}
The DRwD approach computes simply the distance between two senses, but does not look at the meanings of the senses in any depth. The DRwS approach represents a concept in terms of its Wordnet synset\footnote{Synset is a set of cognitive synonyms.}, synonyms, hyponyms and hypernyms. This approach is like the DRwD approach, but instead of computing the distance between lexical entries in each pair, we calculate the similarity, called {\em simValue}. If the {\em simValue} of a {\em <$LexicalEntry_i$,$LexicalEntry_j$>, $i\neq j$} pair is equal or larger than $\beta$, we conclude that the {\em $LexicalEntry_i$} has the same meaning as {\em $LexicalEntry_j$}.

To calculate  \emph{simValue} between two phrases, we obtain the \emph{ExpansionSet} for every word in each phrase from the WordNet database.  An \emph{ExpansionSet} of a phrase is a union of synset, and/or synonym, and/or hyponym, and/or hypernym of every word in it.  We compare the similarity between the \emph{ExpansionSet}s. The value of $\beta$ and the kinds of \emph{ExpansionSet}s are changed to create different {\em ReverseDictionary}s.  Based on experiments, we find that the best value of $\beta$ is 0.9, and the best \emph{ExpansionSet} is the union of synset, synonyms, hyponyms, and hypernyms. The algorithm for computing the {\em simValue} of entries is shown in Algorithm 3.

\begin{algorithm}[h!]
\caption{simValue($LexicalEntry_i$, $LexicalEntry_j$)}
\begin{algorithmic}
\STATE $simWords$ := $\phi$
\IF {$LexicalEntry_i.LexicalUnit$ \& {\em$LexicalEntry_j.LexicalUnit$} have a Wordnet lexical ontology}
\FORALL {(${LexicalUnit}_u \in LexicalEntry_i)$ $\&$ $({LexicalUnit}_v \in LexicalEntry_j$)}
\STATE Find $ExpansionSet$ of every $LexicalEntry$ based on $LexicalUnit$
\ENDFOR
\ELSE 
\FORALL {$({Sense}_u \in LexicalEntry_i)$ $\&$ $({Sense}_v \in LexicalEntry_j)$}
\STATE Find $ExpansionSet$ of every $LexicalEntry$ based on $Sense$
\ENDFOR
\ENDIF
\STATE simWords $\leftarrow$ ExpansionSet ($LexicalEntry_i$) $\cap $ ExpansionSet($LexicalEntry_j$) 
\STATE n $\leftarrow$ExpansionSet($LexicalEntry_i$).length
\STATE m $\leftarrow$ExpansionSet($LexicalEntry_j$).length
\STATE simValue$\leftarrow$min\{$simWords.length\over n$,$simWords.length \over m$\}
\end{algorithmic}
\end{algorithm}

\section{Experimental results}
The goals of our study are to create the high-precision reverse dictionaries, and to increase the numbers of lexical entries in the created dictionaries. 
Evaluations were performed by volunteers who are fluent in both source and destination languages. To achieve  reliable judgment, we use the same set of 100 non-stop word ENG words, randomly chosen from a list of the most common words\footnote{http://www.world-english.org/english500.htm}. We pick randomly 50 words from each  created  $ReverseDictionary$ for  evaluation. Each volunteer was requested to evaluate using a 5-point scale, 5: excellent, 4: good, 3: average, 2: fair, and 1: bad. The average scores of entries in the \emph{ReverseDictionary}s is presented in Figure 1. The DRwS dictionaries are the best in each case. The percentage of agreements between raters is in all cases is around 70\%.

The dictionaries we work with frequently have several meanings for a word. Some of these meanings are unusual, rare or very infrequently used. The DR algorithm creates entries for the rare or unusual meanings by direct reversal. We noticed that our evaluators do not like such entries in the reversed dictionaries and mark them low. This results in lower average scores in the DR algorithm comparing to averages cores in the DRwS algorithm. The DRwS algorithm seems to have removed a number of such unusual or rare meanings (and entries similar to the rare meanings, recursively) improving the average score

Our proposed approaches do not work well for dictionaries containing an abundance of complex phrases. The original dictionaries, except the VIE-ENG dictionary, do not contain many long phrases or complex words. In Vietnamese, most words we find in the dictionary can be considered compound words composed of simpler words put together. However, the component words are separated by space. For example, "{\fontencoding{T5}\selectfont
b\'ai th\`\acircumflex n gi\'ao}" means "idolatry". The component words are "{\fontencoding{T5}\selectfont
b\'ai}" meaning "bow low"; "{\fontencoding{T5}\selectfont th\`\acircumflex n}" meaning "deity"; and "{\fontencoding{T5}\selectfont gi\'ao}" meaning "lance", "spear", "to teach", or "to educate". The presence of a large number of compound words written in this manner causes problems with the ENG-VIE dictionary. If we look closely at Figure 1, all language pairs, except ENG-VIE show substantial improvement in score when we compare the DR algorithm with DRwS algorithm.
 \vspace*{-.055in} 
\begin{figure}[h!]
  \caption{Average entry score in $ReverseDictionary$}
  \centering
    \includegraphics[width=0.5\textwidth]{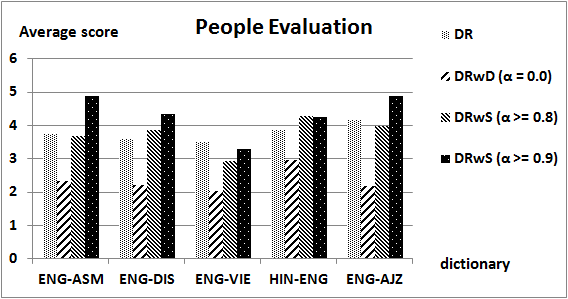}
\end{figure}

The DRwD approach significantly increases the number of entries, but the accuracy of the created dictionaries is much lower. The DRwS approach using  a union of synset, synonyms, hyponyms, and hypernyms of words, and $\beta \geq 0.9$ produces the best reverse dictionaries for each language pair. The DRwS approach increases the number of entries in the created dictionaries compared to the DR algorithm as shown in Figure 2.

\begin{figure}[h!]
  \caption{Number of lexical entries in $ReverseDictionary$s generated from 100 common words}
  \centering
    \includegraphics[width=0.5\textwidth]{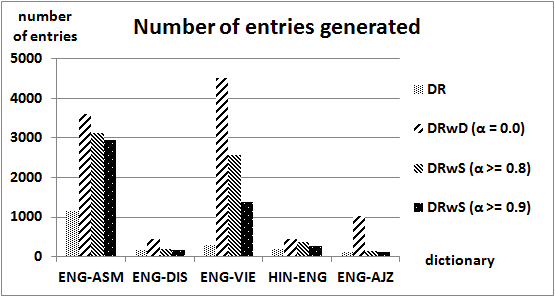}
\end{figure}

We also create the entire reverse dictionary for the AJZ-ENG dictionary. The total number of entries in the ENG-AJZ dictionaries created by using the DR algorithm and DRwS algorithm are 4677 and 5941, respectively. Then, we pick 100 random words from the ENG-AJZ created by using the DRwS algorithm for evaluation. The average score of every entry in this created dictionary is 4.07. Some of the reversal bilingual dictionaries can be downloaded at http://cs.uccs.edu/~linclab/creatingBilingualLexicalResource.html.

\section{Conclusion}
We proposed approaches to create a  reverse dictionary from an existing bilingual dictionary using  Wordnet. We show that a high precision reverse dictionary can be created without using any other intermediate dictionaries or languages. Using the Wordnet hierarchy increases the number of entries in the created dictionaries. We perform experiments with several resource-poor languages including two that are in the UNESCO's list of endangered languages.

\section*{Acknowledgements}
We would like to thank the volunteers evaluating the dictionaries we create: Morningkeey Phangcho, Dharamsing Teron, Navanath Saharia, Arnab Phonglosa, Abhijit Bendale, and Lalit Prithviraj Jain. We also thank all friends in the Xobdo project who provided us with the ASM-ENG-DIS-AJZ dictionaries.

\end{document}